\pdfoutput=1

\documentclass[11pt]{article}

\usepackage{acl}
\usepackage{times}
\usepackage{latexsym}

\usepackage[T1]{fontenc}

\usepackage[utf8]{inputenc}

\usepackage{microtype}

\usepackage{inconsolata}

\usepackage{tikz}
\usepackage{tabularx}
\usepackage{amsmath}
\usepackage{amsthm}
\usepackage{amssymb}
\usepackage{paralist}
\usepackage{multirow}
\usepackage{booktabs}
\usepackage{todonotes}
\usepackage[capitalise,noabbrev]{cleveref}

\usepackage{xspace}

\title{Distilling Robustness into Natural Language Inference Models with Domain-Targeted Augmentation}

\author{Joe Stacey\\
  Imperial College London \\
  \texttt{j.stacey20@imperial.ac.uk}
  \And
  Marek Rei \\
  Imperial College London \\
  \texttt{marek.rei@imperial.ac.uk}
 }

\date{}
\begin{document}
\maketitle

\begin{abstract}
%
Knowledge distillation optimises a smaller student model to behave similarly to a larger teacher model, retaining some of the performance benefits.
While this method can improve results on in-distribution examples, it does not necessarily generalise to out-of-distribution (OOD) settings. 
We investigate two complementary methods for improving the robustness of the resulting student models on OOD domains.
The first approach augments the distillation with generated unlabelled examples that match the target distribution. 
The second method upsamples data points among the training set that are similar to the target distribution.
When applied on the task of natural language inference (NLI), our experiments on MNLI show that distillation with these modifications outperforms previous robustness solutions.
We also find that these methods improve performance on OOD domains even beyond the target domain.\footnote{ \url{https://github.com/joestacey/robust_KD}}
\end{abstract}

\section{Introduction}

Large pre-trained language models can achieve impressive performance across a range of natural language understanding tasks \cite{he2021deberta, LlaMa, GPT3}. However, as performance has increased, so has the number of model parameters \cite{zhao2023survey}. While large models can be impractical for many applications, knowledge distillation can be used to reduce their size \cite{distil_bert, xu2022survey_kd, Gou_2021_kd_survey}.
During distillation, a smaller student model is trained to mimic the behaviour of a more complex teacher model on its training data, often improving the performance of the student model on in-distribution examples. However, this does not necessarily lead to robust improvements that generalise to out-of-distribution settings
\cite{kd_smoothing_robustness, mate-KD, comKD, when_distillation_is_robust}.
We investigate two methods for improving out-of-distribution performance of the resulting student models: 1) Augmenting the distillation by generating new unlabelled task-specific examples that match the target distribution, and 2) Upsampling examples among the training data that are similar to the target distribution. We show that these two approaches are orthogonal and can be effectively combined together.

We use language models (LM) and a multi-step prompting process to generate additional unlabelled examples that target a particular task and domain. 
While labels for these examples can also be generated \cite{nli_gpt3_aug_data}, our experiments show that the LM-generated labels are unreliable and lead to poor performance when used directly for supervised training of the student model.
Instead, we use these examples to gather predicted probability distributions from the teacher models, then optimise the student models to predict similar distributions during distillation. 
This approach overcomes the issue of noisy labels and manages to considerably improve student model performance on out-of-distribution examples.
In contrast to prior work on generating in-domain examples using the training set \cite{mate-KD, comKD, tang_kd_data_aug, haidar2022cilda}, this study is the first to investigate the use of generated data to target specific out-of-distribution domains.

While domain-targeted augmentation of the data improves performance in many settings, we found that it has little effect on minority examples\footnote{A term used to describe examples that counter common spurious patterns in a dataset \cite{tu-etal-2020-empirical}}.
Therefore, we investigate an additional method of upsampling minority examples during distillation, which substantially improves the performance of the student model on adversarial NLI test sets. 
Both of these methods can be combined together to improve robustness across a range of different NLI settings.
While most distillation is performed with a single teacher model, we also experiment with these methods by distilling from an ensemble of models.
Ensembles can be used to better identify minority examples, while also increasing the general robustness of the teacher predictions. To the best of our knowledge, this is the first work to investigate model ensembles for better identification of minority examples.


We evaluate the distillation methods on the task of Natural Language Inference (NLI).
In particular, we aim to improve the robustness of models trained on SNLI \cite{bowman-etal-2015-large} and evaluated on MNLI \cite{williams-etal-2018-broad} (and vice versa) -- a setting where prior work has consistently found negative results or limited improvements \cite{teney2020learning, karimi-mahabadi-etal-2020-end, belinkov-etal-2019-dont, stacey-etal-2020-avoiding, Joe_human_expl, kumar-talukdar-2020-nile, zhao2020lirex}. We find that our simple but novel approach proves to be highly effective, combining the strengths of both LLMs and classification models.

\section{Methods}
Given either a large teacher model or an ensemble of teacher models, we aim to distil these models into a single student model that will perform well on different, out-of-distribution datasets, while also performing well on the in-distribution data used to train the teacher model. In the case of MNLI, our out-of-distribution data consists of multiple, different domains. By generating additional data for some of these domains (our target domains), we can measure how much performance improves on both the target domains and other out-of-distribution data comprised of different domains.

\subsection{Knowledge Distillation}
To maximise in-distribution performance, knowledge distillation often supervises a student model using a combination of the training labels and the soft predictions from a teacher model \cite{hahn2019selfdistil_nlp, kd_smoothing_robustness}. We initially use the training labels, before the student model learns from the teacher model predictions for both the original training data and the augmented data. In effect, we are distilling one fine-tuned model into another fine-tuned model, which we find gives us the best performance. Similar to \citet{comKD}, we consider squared errors for our distillation loss:
\begin{equation}
Loss = \sum_{n=1}^{N}\sum_{c=1}^{C}(p_{n,c} - q_{n, c})^{2}
\end{equation}
where $p_{n,c}$ are the student predicted probabilities for the $c$-th class and $n$-th observation, and $q_{n,c}$ are the corresponding teacher predicted probabilities.

For labelled examples (i.e. not for our augmented data), we only include a distillation loss if either: 1) the teacher predictions are correct, or 2) the teacher model has a larger predicted probability for the correct class compared to the student model. We find that this further improves the performance of the knowledge distillation baseline.

We additionally consider robustness in a self-distillation setting, using the same model architecture for both the student and teacher models \cite{furlanello2018born}. In this case, we experiment with distilling from an ensemble of teacher models rather than using a single teacher model. We consider whether using an ensemble of teacher models improves robustness, and whether our proposed methods are still effective in this setting.
In these cases, the ensemble distillation loss can be described as:
\begin{equation}
Loss = \sum_{n=1}^{N}\sum_{c=1}^{C}(p_{n,c} - \frac{1}{E}\sum_{i=1}^{E}q_{i, n, c})^{2}
\end{equation}
where $p_{n, c}$ are the predicted probabilities from the student model for class $c$ for the $n$-th observation. E represents the total number of models in our ensemble, with $q_{i, n, c}$ representing the predicted probabilities for the $c$-th class from the $i$-th teacher model on the $n$-th observation.

\subsection{Generating Domain-Targeted Data}
For our domain-targeted augmentation (DTA) method,  we consider the MNLI genres as our target domains. Each of these domains is different from the single genre contained within SNLI. To improve performance on these out-of-distribution domains, we generate examples from a GPT-3 model \cite{GPT3} to mimic text that may appear in these genres. To ensure we are testing zero-shot performance and not few-shot performance, we do not provide our generator with any examples from the target genres. Instead, we provide the generator with a high-level description about the genre, and ask the model to generate a premise. For example, for the popular magazine article genre, we use the prompt: `Provide a sentence from a popular magazine article'. We then generate corresponding hypotheses, asking the model to create a hypothesis for each class 
(see \cref{sec:appendix_full_prompts} for our full prompts). While the labels associated with each generated example are unreliable, we use this approach to ensure that our generated examples are relatively balanced across the different classes. This method produces related sentence pairs, with a mixture of entailment, neutral and contradiction relationships, but without a reliable label that we can use during training.

The MNLI-matched and MNLI-mismatched validation sets each consist of five, distinct genres. We generate additional data for 4 of the 5 genres contained within the MNLI-matched validation set (we exclude the telephone transcripts genre). Then, as MNLI-mismatched consists of 5 genres that are not in MNLI-matched, we use MNLI-mismatched to test how well the new data augmentation helps models generalise to new, additional genres. In total, we generate 47,955 unlabelled sentence pairs for the 4 MNLI-matched genres.

\subsection{Augmentation to Address the NLI Word-Overlap Heuristic}

To examine whether using generated, unlabelled data during distillation can also help to address specific, known dataset biases, we also introduce a word-overlap augmentation (WOA). WOA is a variation of our DTA method that aims to overcome the NLI word-overlap bias \cite{mccoy-etal-2019-right}. The word-overlap bias is a heuristic where sentence pairs with a high overlap of words are more likely to be predicted as entailment. HANS measures model performance on this heuristic, containing examples where a high lexical overlap no longer correlates with the entailment class.  

To generate the data, first we ask our generator to provide a short sentence, specifying a conjunction that must be included from a list of 60 conjunctions (ensuring variety in the linguistic structure of our premises). To prevent the model creating a second sentence very similar in meaning, the list of words is then shuffled with the conjunction removed (see Figure \ref{hans_figure}). We use the generator to exclude examples where both sentences have essentially the same meaning, or where the generator finds one of the sentences to be incoherent. In total, 4,695 additional examples were created. 
Similar to our domain-targeted augmentation, the augmented data contains both entailment and non-entailment examples, but no labels are provided.

\begin{figure*}
    \includegraphics[width=460pt]{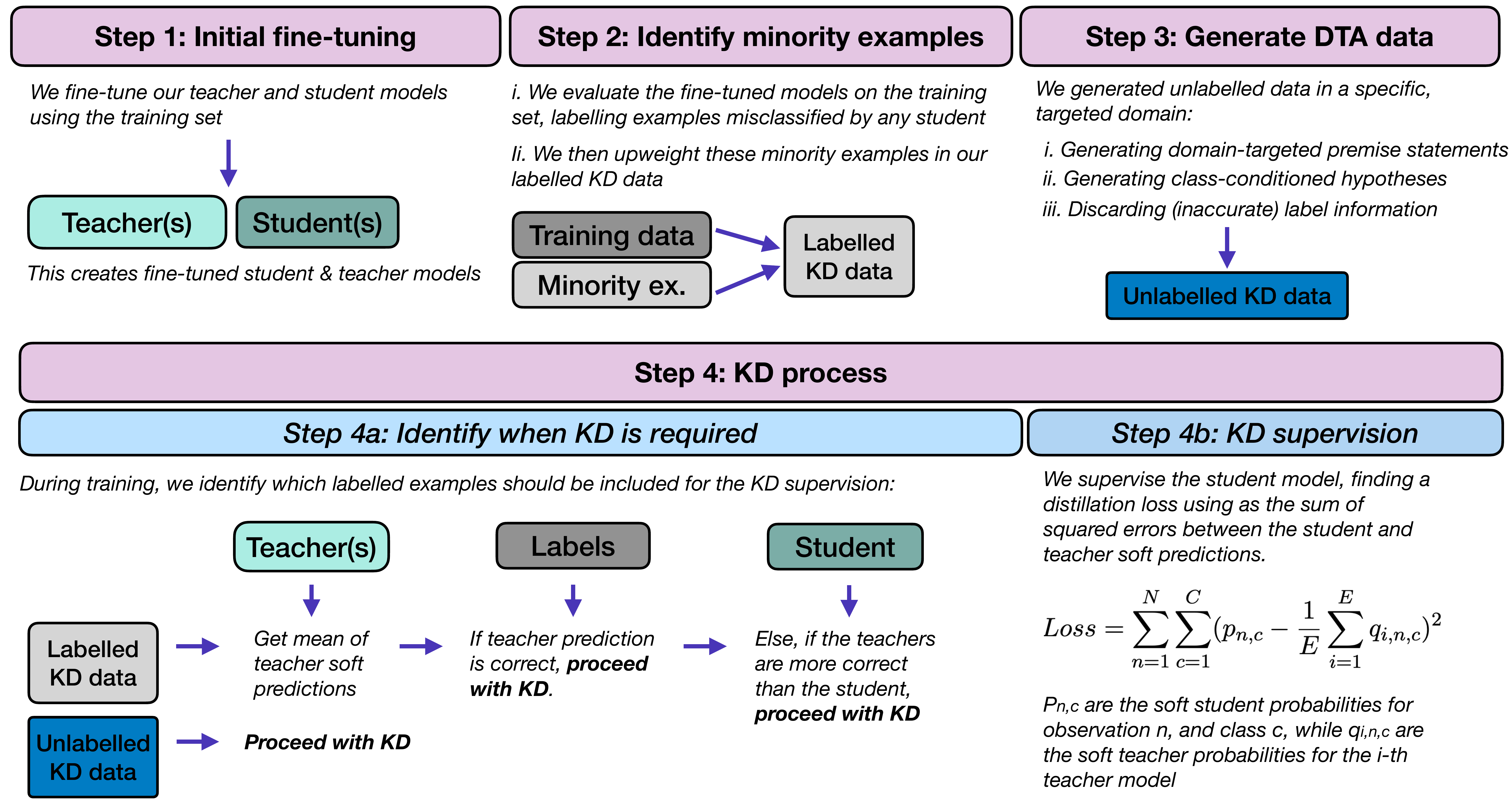} 
    \caption{The full process for applying our DTA and DMU methods together. This diagram includes an explanation of how ensembles can be used in both DTA (with an ensemble of student models) and DMU (with an ensemble of student and/or teacher models).} \label{full_method_diagram}
\end{figure*}

\subsection{Distilled Minority Upsampling}

While our domain-targeted augmentation improves performance on different, unseen domains, it is unlikely to help with in-domain minority examples. We therefore introduce distilled minority upsampling (DMU) as a new method for improving model robustness, upsampling minority examples during knowledge distillation. We are motivated by creating complementary methods that can improve robustness on both minority examples and on different, out-of-distribution domains.

DMU is inspired by Just Train Twice (JTT) \cite{just_train_twice}, which \citet{kd_smoothing_robustness} introduce to a student-teacher setting by identifying examples that a teacher model has misclassified and upsampling these examples when training the student model. Unlike JTT, our DMU method: 1) upsamples the minority examples during distillation rather than during fine-tuning, combining the benefits of both knowledge distillation and the additional supervision for minority examples, and 2) identifies minority examples as observations that the student model, rather than the teacher model, has misclassified. These changes result in a step-change in performance, with DMU substantially outperforming JTT on SNLI-hard \cite{gururangan-etal-2018-annotation}. Additionally, we improve DMU by using an ensemble of models to identify the minority examples, defining minority examples as observations misclassified by any model in the ensemble. 

A summary of how both DTA and DMU are applied together is provided in \cref{full_method_diagram}.
\section{Experiments}
\subsection{Distillation Setup}
We experiment extensively with our new robustness methods across different distillation settings. While generative language models have shown excellent performance on a wide range of tasks, recent work has found that they still underperform on NLI compared to discriminative models \citep{gpt3-not-robust,wei2023chainofthought}. Therefore, we use generative models only for example generation and fine-tune pre-trained discriminative models for the NLI classification task. We evaluate the following combinations of teacher and student models: 1) a TinyBERT \cite{jiao2020tinybert} student model and a BERT \cite{devlin-etal-2019-bert} teacher model, 2) a BERT student model and a BERT teacher model, 3) a DeBERTa \cite{he2021deberta} student model and a DeBERTa teacher model, 4) a BERT student model and a DeBERTa teacher model and 5) a TinyBERT student model and a DeBERTa teacher model. In settings 2 and 3 (using self-distillation), we experiment with using an ensemble of teacher models. For each result, we report an average from 10 different seeds, performing significance testing\footnote{We use two-tailed bootstrapping hypothesis testing \citep{efron1993introduction} to test statistical significance.} in each case.

We additionally experiment with applying DMU when distilling a RoBERTa-large \cite{liu2019roberta} teacher model into a distil-RoBERTa student model, providing a comparison with previous work. This involves training on MNLI, and testing out-of-distribution performance on HANS. We additionally test our word-overlap augmentation method in this same setting. 

\subsection{Domain-Targeted Augmentation}
When applying our domain-targeted augmentation (DTA) method, we primarily use SNLI as our in-distribution data and MNLI as our out-of-distribution data, using our generated examples for MNLI as unlabelled data during the distillation. We test out-of-distribution performance on both MNLI-matched and MNLI-mismatched, despite only generating data for the genres contained within MNLI-matched. We perform additional experimentation using MNLI as the training data and SNLI as the out-of-distribution data. The data is generated for SNLI using the same process as described for MNLI, with 47,898 unlabelled examples created for SNLI. 
\begin{table*}[!t]
\begin{center}
\begin{tabular}{rcccccc}
\toprule
\multirow{2}{*}{\bf Model} & \multicolumn{2}{c}{\bf In-Distribution} & \multicolumn{3}{c}{\bf Out-of-Distribution} \\
\cmidrule(lr){2-3} \cmidrule(lr){4-6}
  &  SNLI-dev & SNLI-test & SNLI-hard & MNLI-mm & MNLI-m \\
\midrule
& \multicolumn{5}{c}{\textit{BERT -> TinyBERT:}} \\
\midrule
BERT teacher & 91.03 & 90.59 & 80.31 & 75.01 & 74.97 \\
\midrule
TinyBERT baseline  & 77.99 & 78.25 & 56.98  & 55.50 & 54.24 \\
Baseline w/ labelled aug. data & 77.72 & 78.18 & 56.73 & 46.47 & 45.55 \\
JTT\textsuperscript{1,2} & 76.96 & 76.25 & 55.93 & 52.66 & 52.00 \\ 
KD (standard distillation) & 80.11 & 80.34 & 60.02 & 57.69 & 55.82 \\ 
KD + Smoothing\textsuperscript{1} & 80.09 & 80.33 & 60.07 & 57.71 & 55.83 \\ 
\midrule
\textit{Ours:} \\
DMU & 80.00 & 80.25$\downarrow$ & 65.94$\uparrow$ & 55.86$\downarrow$ & 54.08$\downarrow$ \\ 
DTA & \textbf{80.16} & \textbf{80.51}$\uparrow$ & 60.26$\uparrow$ & \textbf{59.94}$\uparrow$ & \textbf{57.17}$\uparrow$ \\ 
DTA with DMU & 80.11 & 80.43$\uparrow$ & \textbf{66.04}$\uparrow$ & 59.01$\uparrow$ & 56.54$\uparrow$ \\ 
\bottomrule
\end{tabular}

\end{center}
\caption{Accuracy of a TinyBERT model (4.4 million parameters), compared to a BERT model (110 million parameters) distilled into a TinyBERT model. We compare performance of standard knowledge distillation to our approach using domain-targeted data augmentation and our DMU approach. We also compare with KD with smoothing \cite{kd_smoothing_robustness}\textsuperscript{1} and a JTT baseline \cite{just_train_twice, kd_smoothing_robustness}\textsuperscript{1,2}. MNLI-m and MNLI-mm refer to the MNLI matched and mismatched validation sets respectively. All distillation and DMU results are an average from 10 seeds. $\uparrow$ and $\downarrow$ represent  statistically significant results ($p<0.05$), with all p-values displayed in \cref{compression_results_p_vals}. The best results are in bold.}
\label{compression_results}
\end{table*}
To show the effect of our domain-targeted augmentation, we compare our results to standard knowledge distillation only using the training data. We also provide a baseline that uses the augmented data as labelled data during training, using the label that the hypotheses were conditioned over during the data generation. An additional smoothing baseline is also provided, as proposed by \citet{kd_smoothing_robustness}, which raises each class prediction by the teacher to the power of 0.9 before normalizing these scores. 


\subsection{Distilled Minority Upsampling}

As DMU provides additional supervision to minority examples which counter common spurious correlations, this is likely to improve performance on minority examples rather than on other unseen out-of-distribution datasets. We therefore train our model on SNLI and evaluate performance on SNLI-hard \cite{gururangan-etal-2018-annotation}, a test set that has been widely used to test robustness \cite{karimi-mahabadi-etal-2020-end, belinkov-etal-2019-dont, belinkov-etal-2019-adversarial, sanh2020learning}. Additionally, we experiment with using an ensemble of models to better identify the minority examples, before performing the distillation using an ensemble of teacher models. We also compare DMU with a JTT baseline \cite{kd_smoothing_robustness}.

\section{Results}

\subsection{Domain-Targeted Augmentation}
Our domain-targeted augmentation (DTA) significantly improves performance on the out-of-distribution MNLI-matched dataset for every condition we tested. In the case where a BERT model is distilled into a TinyBERT model, out-of-distribution performance on MNLI-matched is +1.35\% higher when compared to applying a knowledge distillation baseline (Table \ref{compression_results}). When distilling a DeBERTa teacher model into either a BERT or TinyBERT student model, we see improvements of 1.8\% and 1.61\% percentage points respectively (see \cref{DeBERTaToBERT} and \cref{compression_results_deberta}).

Not only does the augmented data improve performance on MNLI-matched in the targeted domains, but also on MNLI-mismatched, which consists of different domains. We observe statistically significant improvements on MNLI-mismatched for every combination of teacher and student models that we tested (see 
\cref{compression_results}, \cref{self_distil}, \cref{DeBERTaToBERT} and \cref{compression_results_deberta}), showing that our method can also improve performance on domains that were not included in the augmented data. While not the focus of our work, we also observe a very small but statistically significant improvement on the in-distribution SNLI-test set in each case. Finally, we see similar improvements when training on MNLI and testing on SNLI (see \cref{train_mnli_test_snli}).

If the augmented data is used as labelled data (using the label the generated hypothesis was conditioned on), out-of-distribution performance on MNLI is substantially worse than the baseline. Interestingly however, the inclusion of this labelled data has little impact in-distribution. 
\begin{table}[!t]
\begin{center}
\begin{tabular}{rccc}
\toprule
 & \multicolumn{2}{c}{\bf Out-of-Distribution} \\
  & MNLI-mm & MNLI-m \\
\midrule
& \multicolumn{2}{c}{\textit{BERT -> BERT:}} \\

\midrule
BERT baseline & 75.01 & 74.97 \\
\midrule
KD & 75.42 & 75.50 \\ 
DTA  (Ours) & 75.77$\uparrow$ & 75.86$\uparrow$ \\ 
\midrule
KD$_{\text{ens}}$ (Ours)& 75.90 & 75.98 \\ 
DTA$_{\text{ens}}$ (Ours) & \textbf{76.42}$\uparrow$ & \textbf{76.45}$\uparrow$ \\ 
\midrule
& \multicolumn{2}{c}{\textit{DeBERTa -> DeBERTa:}} \\
\midrule
DeBERTa baseline & 84.78 & 84.56 \\
\midrule
KD & 85.24 & 84.83 \\ 
DTA (Ours) & 85.68$\uparrow$ & 85.20$\uparrow$ \\ 
\midrule
KD$_{\text{ens}}$ (Ours) & 85.52 & 85.29 \\ 
DTA$_{\text{ens}}$ (Ours)& \textbf{86.18}$\uparrow$ & \textbf{85.77}$\uparrow$ \\
\bottomrule
\end{tabular}

\end{center}
\caption{DTA is tested in a self-distillation setting, using either a single teacher or an ensemble of teachers. All distillation results are an average from 10 seeds. $\uparrow$ and $\downarrow$ represent statistically significant results ($p<0.05$), testing the significance of DTA compared to standard knowledge distillation. The best results are in bold.}
\label{self_distil}
\end{table}
\begin{table}[!t]
\begin{center}
\begin{tabular}{rccc}
\toprule

\textbf{Method} & \textbf{Test} & \textbf{SNLI-Hard} & $\Delta$ \\
\midrule
Baseline & 78.25 & 56.98 & \\
JTT & 76.25 & 55.93 & -1.05 \\
KD & 80.34 & 60.02 & +3.04 \\
\midrule
DMU & 80.25$\downarrow$ & 65.94$\uparrow$ & +8.96 \\
DMU$_{\text{up}}$ & 80.88$\uparrow$ & 66.42$\uparrow$ & +9.44 \\
DMU$_{\text{full}}$ & \textbf{81.01}$\uparrow$ & \textbf{66.48}$\uparrow$ & +9.50 \\

\bottomrule
\end{tabular}

\end{center}
\caption{Performance of JTT \cite{just_train_twice, kd_smoothing_robustness} compared DMU using a TinyBERT student and BERT teacher model. DMU$_{\text{up}}$ uses a single teacher model but upsamples examples that any model in an ensemble incorrectly predicted, while DMU$_{\text{full}}$ also uses an ensemble of teachers during distillation. All DMU results are an average from 10 seeds. $\uparrow$ and $\downarrow$ represent results that are statistically significant results ($p<0.05$).}
\label{jtt_table}
\end{table}

\subsection{Distilled Minority Upsampling}
For each condition we tested, DMU significantly improves performance on the SNLI-hard test set (see 
\cref{compression_results}, \cref{DeBERTaToBERT} and \cref{compression_results_deberta}). When using a BERT teacher and TinyBERT student model, this improvement is substantial, with 5.92\% higher accuracy than the knowledge distillation baseline (Table \ref{compression_results}). This improvement on SNLI-hard contrasts with the results from DTA, which does not substantially improve performance on these minority examples.

While previous work using JTT in a teacher-student setting uses the teacher model to identify minority examples \cite{kd_smoothing_robustness}, we find better performance when using the student model to identify minority examples (see Appendix Table \ref{jtt_table_teacherminority} for results using teacher-identified minority examples).

DMU and DTA are complementary, and when applying both methods we see statistically significant improvements on MNLI-matched, MNLI-mismatched and SNLI-hard in every condition tested (see \cref{compression_results}, \cref{DeBERTaToBERT} and \cref{compression_results_deberta}). While including DMU with DTA can reduce performance on MNLI compared to only using DTA (see \cref{compression_results}), there are corresponding substantial improvements on SNLI-hard. On the other hand, including DTA with DMU mitigates some of the limitations of using DMU, which can otherwise have reduced performance on MNLI-matched and MNLI-mismatched, or reduced performance in-distribution (see 
\cref{compression_results} and \cref{compression_results_deberta}).

\subsection{Distilling from an Ensemble of Teacher Models}

Performing distillation with an ensemble of teacher models has significantly better performance, both in-distribution and out-of-distribution, compared to distillation with a single teacher model. This is the case for both our BERT and DeBERTa models (see \cref{ens_benefit_table}). Additionally, we find that our domain-targeted data augmentation significantly improves performance when combined with an ensemble of teacher models. This is the case when using either a BERT or DeBERTa model as the student and teacher (see \cref{self_distil} and \cref{self_distil_all}). While these improvements are small, in the case of BERT including our augmented data with the ensemble improves performance on  MNLI matched and mismatched by 58\% and 47\% relative to the baseline. 


We also find that we can improve DMU by using an ensemble to identify minority examples, upsampling examples that have been incorrectly predicted by any model in an ensemble (DMU$_{\text{up}}$ in Table \ref{jtt_table}). In this case, DMU$_{\text{up}}$ improves performance by 9.44\% compared to the baseline student model. This ensemble consists of the student model, and 7 other models consisting of the same architecture. Additionally, instead of using a single teacher model (DMU$_{\text{up}}$), an ensemble of teacher models can also be used during the distillation (DMU$_{\text{full}}$ in Table \ref{jtt_table}), slightly improving performance.

\subsection{Improving Robustness against the Word-Overlap Heuristic}

We find that our word-overlap augmentation (WOA) improves performance on the adversarial HANS dataset after training on MNLI (Table \ref{train_mnli_test_hans}), although without setting a new state-of-the-art result. Previous work augments the training data with a large number of training examples (392,702, the same size as the MNLI training set) \cite{comKD}, whereas we only augment the data with 4,695 examples which we upsample (x10). When we include our small number of domain-targeted observations, we achieve more than half the out-of-distribution improvements compared to the prior work while only using a fraction (1.2\%) of the additional examples. Directly applying our DMU method proves to be highly effective in this setting, outperforming previous work without any data augmentation.



\begin{table}[!t]
\begin{center}
\begin{tabular}{rccc}
\toprule

\textbf{Method} & \textbf{MNLI-m} & \textbf{HANS} & \textbf{\#aug} \\
\midrule
\multicolumn{4}{l}{\textit{Teacher and student models}} \\
\midrule
RoBERTa-large & 89.6 & 76.6 \\
DistilRoBERTa & 83.8 & 59.9 \\
\midrule
\multicolumn{4}{l}{\textit{Without augmentation}} \\
\midrule
Annealing-KD\textsuperscript{1} & \textbf{84.5} & 61.2 & - \\
KD & 84.1 & 61.8 \\
DMU$_{\text{full}}$ & 84.2 & \textbf{65.9} & - \\
\midrule
\multicolumn{4}{l}{\textit{With augmentation}} \\
\midrule
ComKD\textsuperscript{2} & \textbf{87.2} & \textbf{68.6}  & 393k \\
WOA$_{\text{ens}}$ & 84.3 & 65.1 & 5k \\
WOA**$_{\text{ens}}$ & 81.6 & 68.3 & 5k \\
\bottomrule
\end{tabular}

\end{center}
\caption{Accuracy is displayed on MNLI-matched (in-distribution), and HANS (out-of-distribution). ** refers to the setting where we only perform the distillation step on the augmented data. We compare our results to previous sota results improving robustness for knowledge distillation: \textsuperscript{1} \citet{jafari2021annealing_kd}, and \textsuperscript{2} \citet{comKD}. The best results are in bold.}
\label{train_mnli_test_hans}
\end{table}

\begin{table*}[!t]
\begin{center}
\begin{tabular}{rllclclc}
\toprule
& & \multicolumn{2}{c}{\textbf{MNLI-mm}} & \multicolumn{2}{c}{\textbf{MNLI-m}} & \multicolumn{2}{c}{\textbf{MNLI-all}} \\
\textbf{Method} & \textbf{Baseline} & \textbf{Acc} & \textbf{Imp} & \textbf{Acc} & \textbf{Imp} & \textbf{Acc} & \textbf{Imp}\\
\midrule
\multicolumn{3}{l}{\textit{Hyper-parameter tuning on MNLI}} \\
\midrule
Negative sampling\textsuperscript{1} & $_{\text{LSTM}}$ & 43.66 & -3.91 & 43.76 & -2.10 & 43.71 & -3.01 \\
Hyp-only adversary\textsuperscript{1} & $_{\text{LSTM}}$ & 49.24 & \textbf{+1.67} & 47.24 & \textbf{+1.38} & 48.24 & \textbf{+1.52} \\
Ensemble-adversaries\textsuperscript{2} & $_{\text{LSTM}}$ &  52.81 & -0.10 & 54.18 & +0.80 & 53.49 & +0.35 \\
Product of Experts\textsuperscript{3} & $_{\text{BERT}}$ & 73.49 & -0.49 & 73.61 & -0.79 & 73.55 & -0.64 \\
Debiased Focal Loss\textsuperscript{3} & $_{\text{BERT}}$ & 74.00 & +0.02 & 73.58 & -0.82 & 73.79 & -0.40 \\
\midrule
\multicolumn{3}{l}{\textit{No hyper-parameter tuning on MNLI}} \\
\midrule
Rationale supervision\textsuperscript{4} & $_{\text{BERT}}$ & 73.36 & +0.84 & 73.19 & +0.91 & 73.28 & +0.87 \\
KD\textsuperscript{ } & $_{\text{BERT}}$ & 75.42 & +0.41 & 75.50 & +0.53 & 75.46 & +0.47 \\
NILE\textsuperscript{5} & $_{\text{RoB.}}$ & 77.22 & -2.07 & 77.07 & -2.22 & 77.15 & -2.14 \\
LIREx\textsuperscript{6} & $_{\text{RoB.}}$ & 79.79 & +0.06 & 79.85 & -0.27 & 79.82 & -0.10 \\
KD\textsuperscript{ } & $_{\text{DeB.}}$ & 85.24 & +0.46 & 84.83 & +0.27 & 85.04 & +0.37 \\
\midrule
\textit{Ours:} \\
DTA$_{\text{ens}}$\textsuperscript{ } & $_{\text{BERT}}$ & 76.42 & \textbf{+1.41} & 76.45 & \textbf{+1.48} & 76.43 & \textbf{+1.44} \\
DTA$_{\text{ens}}$\textsuperscript{ } & $_{\text{DeB.}}$ & \textbf{86.18} & +1.40 & \textbf{85.77} & +1.21 &  \textbf{85.98} & +1.31 \\
\bottomrule
\end{tabular}
\end{center}
\caption{A comparison of work testing zero-shot performance on the MNLI matched and mismatched sets after training on SNLI (MNLI-all combines both validation sets). Performance of each method is compared to their respective baselines to show when further out-of-distribution improvements are achieved. RoB. stands for RoBERTa, while DeB stands for DeBERTa. \textsuperscript{1} \citet{belinkov-etal-2019-dont}, \textsuperscript{2} \citet{stacey-etal-2020-avoiding}, 
\textsuperscript{3} \citet{karimi-mahabadi-etal-2020-end},
\textsuperscript{4} \citet{Joe_human_expl}, \textsuperscript{5} \citet{kumar-talukdar-2020-nile}, \textsuperscript{6} \citet{zhao2020lirex}. Methods 4, 5 \& 6 include the use of human annotated rationales \cite{camburu2018esnli}.}
\label{mnli_ood_comparisons}
\end{table*}

\subsection{Comparison to Previous OOD-Performance on MNLI}

Improving performance out-of-distribution on MNLI after training on SNLI remains a challenging task. Despite extensive prior work evaluating models in this condition, few approaches yield out-of-distribution improvements. We compare this prior work to our own results using self-distillation with BERT and DeBERTa. We find that distillation using both our domain-targeted augmentation and using an ensemble of teachers outperforms all previous work (Table \ref{mnli_ood_comparisons}). While adversarial training using a single hypothesis-only adversary \cite{belinkov-etal-2019-dont} produced larger improvements, this work involved hyper-parameter tuning on MNLI-mismatched for a model evaluated on MNLI-matched, and vice versa. On the other hand, our experiments also do not assume the availability of any MNLI examples to use as a validation set. 
Previous related work includes debiasing techniques that aim to improve zero-shot performance \cite{belinkov-etal-2019-dont, stacey-etal-2020-avoiding, karimi-mahabadi-etal-2020-end, teney2020learning}, in addition to previous work incorporating human explanations when training \cite{zhao2020lirex, kumar-talukdar-2020-nile, Joe_human_expl}.

\section{Related Work}

\subsection{Improving Robustness in Knowledge Distillation}
To improve robustness in knowledge distillation, previous methods have involved smoothing the teacher predictions \cite{kd_smoothing_robustness, jafari2021annealing_kd}, or using additional unlabelled training data during the distillation \cite{mate-KD, comKD}. The smoothing methods either smooth the teacher model predictions more at the beginning of training \cite{jafari2021annealing_kd}, or based on the difficulty of each example \cite{kd_smoothing_robustness}. We find that our results outperform a smoothing baseline proposed by \citet{kd_smoothing_robustness}.

Most similar to our approach of using additional, unlabelled data during distillation, \citet{mate-KD, comKD,haidar2022cilda} augment their model with additional training examples that are created by perturbing existing observations. This involves randomly masking words, before replacing these words using a generator that is trained to maximise the difference between the student and teacher predicted probabilities \cite{mate-KD, comKD, haidar2022cilda} and their intermediate representations \cite{haidar2022cilda}. Further work involves perturbing existing examples without an adversarial objective \cite{tang_kd_data_aug, jiao2020tinybert}, and perturbing additional language data not related to NLI into additional NLI examples \cite{rashid2020zeroshot}. Previous work measures adversarial robustness using HANS \cite{comKD, mate-KD, kd_smoothing_robustness, haidar2022cilda}. We compare our DMU and WOA methods to \citet{comKD} and \citet{jafari2021annealing_kd}, the state-of-the-art results for robust knowledge distillation on HANS with and without additional data augmentation.


\subsection{Upsampling Minority Examples}

\citet{tu-etal-2020-empirical} introduce 
the term minority examples to describe instances which counter the spurious correlations present in a dataset. Upsampling these minority examples during training has been shown to improve model robustness \cite{just_train_twice, yaghoobzadeh_minority_examples, kd_smoothing_robustness}. While \citet{just_train_twice} identify minority examples as training examples that are misclassified by a model trained on that data, \citet{yaghoobzadeh_minority_examples} additionally consider instances that have been properly classified at some point during training, but are then misclassified later in training. Rather than upsampling minority examples, \citet{korakakis-vlachos-2023-improving} introduce a minimax objective to improve robustness, upweighting the loss of examples during training to maximise the training loss. \citet{kd_smoothing_robustness} adapt these ideas to a student-teacher setting, upsampling training examples that a teacher model has misclassified when training a student model. We directly compare our DMU method to this approach, showing substantial improvements in robustness.

\subsection{Language Model Data Augmentation for Knowledge Distillation}

Using language models to generate additional data has previously shown promising results in a data-free setting. \citet{data_free_kd} generate synthetic examples based on the topics present in the training data, using the generated data to perform knowledge distillation in a data-free setting. 

Alternatively, without being provided with a more specific prompt, language models can be fine tuned on the training data to generate additional training examples that can be used with the distillation \cite{he2022generate}. Language models can also be used to modify spans in NLI examples, creating new counterfactual examples \cite{chen2022disco}, before an NLI model decides whether the perturbed counterfactual examples have the desired class. Rather than creating counterfactual training data, or perturbing existing training examples, we generate data that specifically targets new, additional domains.

\subsection{Knowledge Distillation with Ensembles}

Knowledge distillation is most commonly applied to distil a single, more complex teacher model into a smaller student model with fewer parameters \cite{He2022_survey_kd, kd_application, Gou_2021_kd_survey}. However, ensembles of models can also be distilled into a single model \cite{hinton2015distilling, asif2020ensemble_kd, kd_ensemble_machine_translation}. We find that the in-distribution improvements from using an ensemble of teacher models are accompanied by out-of-distribution improvements. Moreover, using these ensembles are complementary to our domain-targeted data augmentation method.

\section{Conclusion}
We introduce domain-targeted augmentation and DMU as two methods to improve out-of-distribution robustness in NLI. In the case of the domain-targeted augmentation, using the additional, generated examples during knowledge distillation proves to be a highly effective technique, outperforming all previous work that measures out-of-distribution robustness on MNLI. Not only do we find that performance is better on the targeted domains, but performance is also better for domains that were not included in the augmented data. We also find that our DMU method produces substantial improvements on SNLI-hard, helping the student model to make better predictions for minority examples. Using ensembles can help both methods, improving how minority examples are identified for DMU, and improving the teacher distributions for our domain-targeted augmentation. 
%

We also find that our WOA method can improve robustness on HANS, showing that using unlabelled data during distillation can also target specific, known dataset biases. 

\section*{Limitations}
The main limitation of our domain-targeted data augmentation is the cost of generating the unlabelled examples using GPT3 (approximately 100USD for the experiments provided). As a result, while we perform extensive experimentation on NLI datasets (with over 200 experiments), we do not also apply this method to other NLP tasks. We choose to focus on NLI, as many previous works on robustness evaluate on this task. Our experimentation is also limited to single sentence NLI datasets such as MNLI, SNLI and HANS, and therefore the findings may not necessarily generalise to NLI datasets with longer hypotheses and premises such as ANLI \cite{nie-etal-2020-adversarial} or ConTRoL \cite{liu2020natural_control}.

In this work we show that including the domain-targeted augmentation benefits other domains that the data was not generated for. We demonstrate this by creating data to mimic the domains within MNLI-matched, before testing performance on MNLI-mismatched which contains a different set of domains. However, as there are similarities between examples in MNLI matched and mismatched, further work could test the extent that these benefits generalise to different tasks or domains.

Additionally, in Table \ref{mnli_ood_comparisons}, we provide all results known to us from methods that train on SNLI and test zero-shot performance on MNLI. While this previous work contains a variety of methods, including different debiasing techniques, not all NLI debiasing methods have been evaluated in this setting.

Finally, for our DTA method it is possible that our GPT-3 generator model has seen data from the target domain during pretraining. However, on inspection, the unlabelled examples generated by GPT-3 did not closely resemble the data in MNLI. This is likely because we generated the data in several stages, first asking the model for an ‘example extract from [target domain]’ to create a premise statement. This prompt refers to a broad topic and is very unlikely to result in GPT-3 generating premises specifically from MNLI. The second sentence (the hypothesis) is generated to be relevant specifically to a given premise - if the premise is not from MNLI then the hypothesis would not be either.

\section*{Ethics Statement}
The data generation process in this paper did not involve any human annotation, and the generated data does not contain personal information. All data is generated using GPT-3 (please see \cref{sec:appendix_exp_setup} for further information).

\section*{Acknowledgements}

We would like to thank He He for all her valuable feedback on this work. Joe Stacey was supported by the Apple Scholars in AI/ML PhD fellowship.

\bibliography{anthology, custom}
\appendix
\section{Additional results}\label{sec:appendix_additional_results}
We provide additional results from training the student and teacher models on MNLI, and testing zero-shot performance on SNLI (Table \ref{train_mnli_test_snli}). For these experiments, we use a BERT teacher model and TinyBERT student model. We find that these results mirror the results from training on SNLI and testing on MNLI (see Table \ref{compression_results}), with 2.01\% and 1.96\% point improvements on the validation and test sets compared to applying knowledge distillation without the unlabelled augmented data. For these experiments, we use the mismatched validation set as our validation set for early stopping.

We also provide experimental results when using a DeBERTa model as a teacher model, and either a BERT model or a TinyBERT model as the student (see Table \ref{DeBERTaToBERT} and Table \ref{compression_results_deberta}). In these settings, we also see improvements from our domain-targeted augmentation, with improvements of 1.84\% on MNLI-mismatched, and improvements of 1.8\% on MNLI-matched for a BERT student model, compared to improvements of 2.4\% and 1.61\% with a TinyBERT student model. When applying our DMU method with a DeBERTa teacher model, we also see statistically significant improvements on SNLI-hard (see \cref{DeBERTaToBERT} and \cref{compression_results_deberta}).

Additionally, we test our DMU method when using the teacher models to identify minority examples instead of using the student models. In this case, we use either a single teacher model, or an ensemble of teacher models to identify these examples. Our results show worse performance when using a teacher model to identify minority examples, compared to when using a student model (see Table \ref{jtt_table_teacherminority}).

\section{Performance of LLMs on NLI}\label{sec:LLMs_NLI}

There is currently a lack of evidence that LLMs outperform other transformer-based classification models on NLI, especially considering the number of model parameters. While LLMs have shown increasingly impressive performance across a range of different tasks, this is not the case with NLI. \citet{gpt3-not-robust} find significant robustness degradation on NLI when using GPT-3.5-turbo, despite finding better performance on other tasks. This work includes an evaluation of GPT-3.5-turbo on both SNLI and MNLI \citep{gpt3-not-robust}.

Similar findings have been found across other NLI datasets. For example, \citet{wei2023chainofthought} apply a Chain-of-Thought GPT-3.5-turbo model, with performance substantially below previous work using a DeBERTa-v3 baseline \citep{joe_anli}.

\section{Robustness in NLI}

Improving model robustness for Natural Language Inference (NLI) is a well studied area, where robustness is measured either by testing performance on adversarial datasets such as HANS or the NLI stress tests \cite{naik-etal-2018-stress}. Alternatively, robustness is measured using unseen, out-of-distribution test sets such MNLI \cite{williams-etal-2018-broad}, a challenging robustness setting where existing de-biasing methods often do not improve performance \cite{belinkov-etal-2019-dont, karimi-mahabadi-etal-2020-end}. There has been some success improving out-of-distribution performance on MNLI \cite{Joe_human_expl, teney2020learning, belinkov-etal-2019-dont}, particularly in a reduced-data setting \cite{Joe_Logic, mahabadi2021variational}, however, most methods do not lead to any improvements \cite{zhao2020lirex, kumar-talukdar-2020-nile, camburu2018esnli, belinkov-etal-2019-dont, karimi-mahabadi-etal-2020-end}. 



\section{Model Parameters}

Our DeBERTa model consists of 184 million parameters, compared to 110 million parameters for BERT and 4.4 million parameters for tinyBERT. When distilling RoBERTa, our RoBERTa model consists of 355 million parameters, compared to 83 million for distil-RoBERTa. Over 200 experiments are conducted, consisting of approximately 2500 GPU hours using RTX6000 GPUs.

\begin{table*}[!t]
\begin{center}
\begin{tabular}{rcccccc}
\toprule
\multirow{2}{*}{\bf Model} & \multicolumn{2}{c}{\bf In-Distribution} & \multicolumn{3}{c}{\bf Out-of-Distribution} \\
\cmidrule(lr){2-3} \cmidrule(lr){4-6}
  &  SNLI-dev & SNLI-test & SNLI-hard & MNLI-mm & MNLI-m \\
\midrule
& \multicolumn{5}{c}{\textit{DeBERTa -> BERT:}} \\
\midrule
DeBERTa teacher & 93.32 & 92.53 & 84.64 & 84.78 & 84.56 \\
\midrule
BERT baseline  & 91.03 & 90.59 & 80.31  & 75.01 & 74.97 \\
KD & 91.66 & 91.04 & 81.26 & 75.21 & 75.61 \\ 
\midrule
DTA (Ours) & 91.77$\uparrow$ & 91.14$\uparrow$ & 81.42 & \textbf{77.05}$\uparrow$ & \textbf{77.41}$\uparrow$\\ 
DMU (Ours) & 91.77$\uparrow$ & \textbf{91.17}$\uparrow$ & 81.62$\uparrow$ & 75.05 & 75.55 \\
DTA with DMU (Ours) & \textbf{91.84}$\uparrow$ & 91.16$\uparrow$ & \textbf{81.64}$\uparrow$ & 76.72$\uparrow$ & \textbf{77.41}$\uparrow$ \\
\bottomrule
\end{tabular}

\end{center}
\caption{Our domain-targeted augmentation method is compared to a knowledge distillation baseline. These experiments use a DeBERTa teacher model and BERT student model. Results use one random seed. All distillation and DMU results show the accuracy from an average of 10 random seeds. $\uparrow$ and $\downarrow$ represent results that are statistically significant with $p < 0.05$. The best results are in bold.}
\label{DeBERTaToBERT}
\end{table*}

\begin{table*}[!t]
\begin{center}
\begin{tabular}{rcccccc}
\toprule
\multirow{2}{*}{\bf Model} & \multicolumn{2}{c}{\bf In-Distribution} & \multicolumn{3}{c}{\bf Out-of-Distribution} \\
\cmidrule(lr){2-3} \cmidrule(lr){4-6}
 & MNLI-mm & MNLI-m &  SNLI-dev & SNLI-test & SNLI-hard \\
\midrule
& \multicolumn{5}{c}{\textit{BERT -> TinyBERT:}} \\
\midrule
BERT teacher & 84.64 & 84.38 & 79.31 & 80.09 & 71.3 \\
\midrule
TinyBERT baseline  & 65.62 & 63.89 & 52.77 & 52.59 & 42.90 \\
KD & 68.23 & 66.72 & 55.78 & 55.90 & 46.15 \\ 
DTA (Ours) & \textbf{68.45}$\uparrow$ & \textbf{66.97}$\uparrow$ & \textbf{57.79}$\uparrow$ & \textbf{57.86}$\uparrow$ & \textbf{46.36} \\ 

\bottomrule
\end{tabular}

\end{center}
\caption{Evaluating our domain-targeted augmentation when training on MNLI and testing zero-shot performance on SNLI. Performance is compared to knowledge distillation without the augmented data, and also a TinyBERT baseline. MNLI-mismatched is our validation set. As SNLI-hard specifically considers minorty examples for models trained on SNLI, we do not also test DMU in this setting. All distillation results show the accuracy from an average of 10 random seeds. $\uparrow$ and $\downarrow$ represent results that are statistically significant with $p < 0.05$. The best results are in bold.}
\label{train_mnli_test_snli}
\end{table*}

\begin{table*}[!t]
\begin{center}
\begin{tabular}{rcccccc}
\toprule
\multirow{2}{*}{\bf Model} & \multicolumn{2}{c}{\bf In-Distribution} & \multicolumn{3}{c}{\bf Out-of-Distribution} \\
\cmidrule(lr){2-3} \cmidrule(lr){4-6}
  &  SNLI-dev & SNLI-test & SNLI-hard & MNLI-mm & MNLI-m \\
\midrule
& \multicolumn{5}{c}{\textit{DeBERTa -> TinyBERT:}} \\
\midrule
DeBERTa teacher & 93.32 & 92.53 & 84.64 & 84.78 & 84.56 \\
\midrule
TinyBERT baseline  & 77.99 & 78.25 & 56.98  & 55.50 & 54.24 \\
KD & 80.00 & 80.24 & 59.85 & 57.67 & 55.90 \\ 
DTA (Ours) & \textbf{80.02} & \textbf{80.40}$\uparrow$ & 60.11$\uparrow$ & \textbf{60.07}$\uparrow$ & \textbf{57.51}$\uparrow$ \\ 
DMU (Ours) & 79.26$\downarrow$ & 79.41$\downarrow$ & 65.39$\uparrow$ & 54.45$\downarrow$ & 53.00$\downarrow$\\
DTA with DMU (Ours) & 79.33$\downarrow$ & 79.65$\downarrow$ & \textbf{65.44}$\uparrow$ & 58.94$\uparrow$ & 56.51$\uparrow$ \\
\bottomrule
\end{tabular}

\end{center}
\caption{Accuracy of a TinyBERT model, compared to a DeBERTa model distilled into a TinyBERT model. We compare performance of standard knowledge distillation to our approach using domain-targeted data augmentation. The best results are in bold. All distillation and DMU results show the accuracy from an average of 10 random seeds. 
$\uparrow$ and $\downarrow$ represent results that are statistically significant with $p < 0.05$. No early stopping was included for our DMU experiments. The best results are in bold.}
\label{compression_results_deberta}
\end{table*}

\begin{table*}[!t]
\begin{center}
\begin{tabular}{rcccccc}
\toprule
\multirow{2}{*}{\bf Model} & \multicolumn{2}{c}{\bf In-Distribution} & \multicolumn{3}{c}{\bf Out-of-Distribution} \\
\cmidrule(lr){2-3} \cmidrule(lr){4-6}
  &  SNLI-dev & SNLI-test & SNLI-hard & MNLI-mm & MNLI-m \\
\midrule
& \multicolumn{5}{c}{\textit{BERT -> BERT:}} \\
\midrule
BERT baseline & 91.03 & 90.59 & 80.31  & 75.01 & 74.97 \\
\midrule
KD & 91.43 & 90.72 & 80.57 & 75.42 & 75.50 \\ 
KD$_{\text{ens}}$  (Ours)& \textbf{91.59}$\uparrow$ & \textbf{90.94}$\uparrow$ & \textbf{80.81}$\uparrow$ & \textbf{75.90}$\uparrow$ & \textbf{75.98}$\uparrow$ \\ 
\midrule
& \multicolumn{5}{c}{\textit{DeBERTa -> DeBERTa:}} \\
\midrule
DeBERTa baseline & 93.32 & 92.53 & 84.64 & 84.78 & 84.56 \\
\midrule
KD & 93.56 & 92.70 & 84.90 & 85.24 & 84.83 \\ 
KD$_{\text{ens}}$ (Ours)& \textbf{93.73}$\uparrow$ & \textbf{92.89}$\uparrow$ & \textbf{85.06} & \textbf{85.52}$\uparrow$ & \textbf{85.29}$\uparrow$ \\ 
\bottomrule
\end{tabular}

\end{center}
\caption{Knowledge distillation for self-distillation is tested for a single teacher model compared to an ensemble of teacher models. All distillation results are an average from 10 random seeds. $\uparrow$ and $\downarrow$ represent results that are statistically significant with $p < 0.05$. The best results are in bold.}
\label{ens_benefit_table}
\end{table*}
\begin{table*}[!t]
\begin{center}
\begin{tabular}{rcccccc}
\toprule
\multirow{2}{*}{\bf Model} & \multicolumn{2}{c}{\bf In-Distribution} & \multicolumn{3}{c}{\bf Out-of-Distribution} \\
\cmidrule(lr){2-3} \cmidrule(lr){4-6}
  &  SNLI-dev & SNLI-test & SNLI-hard & MNLI-mm & MNLI-m \\
\midrule
& \multicolumn{5}{c}{\textit{BERT -> BERT:}} \\
\midrule
BERT baseline & 91.03 & 90.59 & 80.31  & 75.01 & 74.97 \\
\midrule
KD & 91.43 & 90.72 & 80.57 & 75.42 & 75.50 \\ 
DTA  (Ours)& 91.40 & 90.78 & 80.70 & 75.77$\uparrow$ & 75.86$\uparrow$ \\ 
\midrule
KD$_{\text{ens}}$  (Ours)& 91.59 & 90.94 & 80.81 & 75.90 & 75.98 \\ 
DTA$_{\text{ens}}$  (Ours)& \textbf{91.65} & \textbf{91.00} & \textbf{81.00} & \textbf{76.42}$\uparrow$ & \textbf{76.45}$\uparrow$ \\ 
\midrule
& \multicolumn{5}{c}{\textit{DeBERTa -> DeBERTa:}} \\
\midrule
DeBERTa baseline & 93.32 & 92.53 & 84.64 & 84.78 & 84.56 \\
\midrule
KD & 93.56 & 92.70 & 84.90 & 85.24 & 84.83 \\ 
DTA (Ours) & 93.55 & 92.69 & 84.84 & 85.68$\uparrow$ & 85.20$\uparrow$ \\ 
\midrule
KD$_{\text{ens}}$  (Ours)& 93.73 & \textbf{92.89} & \textbf{85.06} & 85.52 & 85.29 \\ 
DTA$_{\text{ens}}$  (Ours)& \textbf{93.74} & 92.82 & 84.97 & \textbf{86.18}$\uparrow$ & \textbf{85.77}$\uparrow$ \\
\bottomrule
\end{tabular}

\end{center}
\caption{As \cref{self_distil} only shows results on MNLI, this table contains self-distillation results from all sets, including SNLI-dev, SNLI-test and SNLI-hard. All distillation results are an average from 10 random seeds. $\uparrow$ and $\downarrow$ represent results that are statistically significant with $p < 0.05$. We test the significance of the domain-targeted augmentation compared to standard knowledge distillation. The best results are in bold.}
\label{self_distil_all}
\end{table*}
\section{Details of experimental setup}\label{sec:appendix_exp_setup}

To generate our domain-targeted data for MNLI, we use a text-curie-001 GPT-3 model to generate both the premises and hypotheses. However, when generating the additional data for HANS, we use text-davinci-003. We use the more expensive davinci model for this setting, as generating sentences that only contained specific words that we provided proved to be a more difficult task for text-curie-001.

We train all baseline models using a learning rate of $10^{-5}$ for 2 epochs using cross entropy loss, with the exception of Distil-RoBERTa (used as the student model in the MNLI-HANS setup). For Distil-RoBERTa and RoBERTa-large, to create a baseline similar to previous work, we train with learning rates of $2\times10^{-5}$ and $5\times10^{-6}$ respectively, in the case of Distil-RoBERTa training for 8 epochs. All baselines are trained with a linear learning rate (increasing for the first half of training, before decreasing for the second half). We use deberta-v3-base for our DeBERTa model, and bert-base-uncased for our BERT model. All baseline models are implemented from HuggingFace \cite{wolf2020huggingfaces}. When using ensembles of BERT models, the eight models we use are different pre-trained models from \citet{sellam2021multiberts} to maximise the variability between each BERT model. 

The distillation stage is performed for 10 epochs with a learning rate of $10^{-6}$, with early stopping applied if there is no improvement within five epochs. The early stopping was not applied when using self-distillation, where we tested using an ensemble of teacher models. In this case, we chose the student model as the model with the best validation performance from the ensemble. Therefore, as the teacher models had lower validation performance than the student, the student validation performance was also likely to decrease during training. We also do not perform early stopping when evaluating on the adversarial HANS dataset, as performance on both MNLI-validation sets are likely to decrease as a result of improvements in HANS, or when distilling a DeBERTa teacher into a TinyBERT student model using DMU, where we do not see improvements in the validation set.

When applying Just Train Twice (JTT) or DMU, the minority examples are upsampled by 6 times, as \citet{just_train_twice} use for MNLI. We also upsample our augmented data for HANS (by 10 times), as we have fewer examples compared to MNLI (4,695 for HANS, compared to 47,955 for MNLI and 47,898 for SNLI). For DMU$_{\text{full}}$, we use an ensemble of 8 models, whereas the self-distillation experiments use an ensemble of 7 models (as one of the 8 models is used as the student model).

\begin{table}[t]
\begin{center}
\begin{tabular}{rccc}
\toprule

\textbf{Method} & \textbf{Test} & \textbf{SNLI-Hard} & $\Delta$ \\
\midrule
Baseline & 78.25 & 56.98 & \\
JTT & 76.25 & 55.93 & -1.05 \\
KD & 80.34 & 60.02 & +3.04 \\
\midrule
\textit{Ours:} \\
DMU$_{\text{teach}}$ & 80.80 & 60.88 & +3.90 \\
DMU$_{\text{t-up}}$ & 80.78 & 61.15 & +4.17 \\
DMU$_{\text{t-full}}$ & \textbf{80.98} & \textbf{61.64} & \textbf{+4.66}\\

\bottomrule
\end{tabular}

\end{center}
\caption{Performance of a JTT baseline \cite{just_train_twice, kd_smoothing_robustness} compared to our DMU$_{\text{teach}}$ method upsampling minority examples during the distillation that a single teacher model has misclassified. We up-sample examples that any model in an ensemble of teacher models incorrectly predicted while still using a single teacher model during the distillation (DMU$_{\text{t-up}}$), or also using an ensemble of teacher models for the distillation (DMU$_{\text{t-full}}$). For DMU$_{\text{t-full}}$, the same teacher models are used to identify the minority examples as those used during the distillation process. The baseline is a TinyBERT student, while JTT and KD methods use a BERT teacher.}
\label{jtt_table_teacherminority}
\end{table}
\section{Full prompts}\label{sec:appendix_full_prompts}

\begin{figure}[ht]
    \includegraphics[width=\columnwidth]{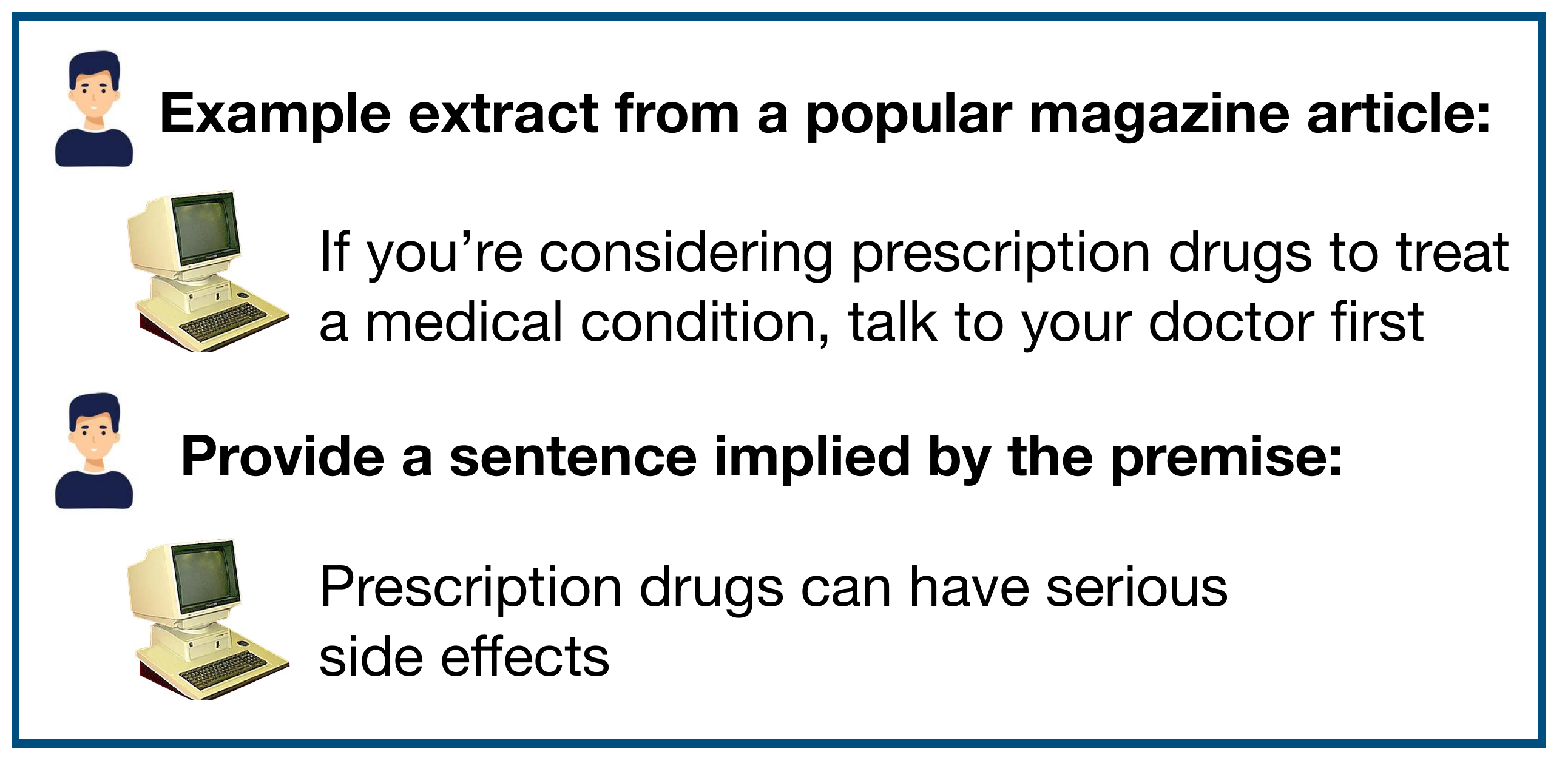} 
    \caption{Our generator model is asked to create a sentence (premise) about a specified genre, before being asked to create a hypothesis that is either implied by the premise, contradicts the premise, or is neutral with respect to the premise. As the hypotheses generated are not faithful to the desired labels (as with this example), we use these examples as unlabelled data during knowledge distillation.} \label{method_example}
\end{figure}

\begin{figure*}[ht]
    \includegraphics[width=450pt]{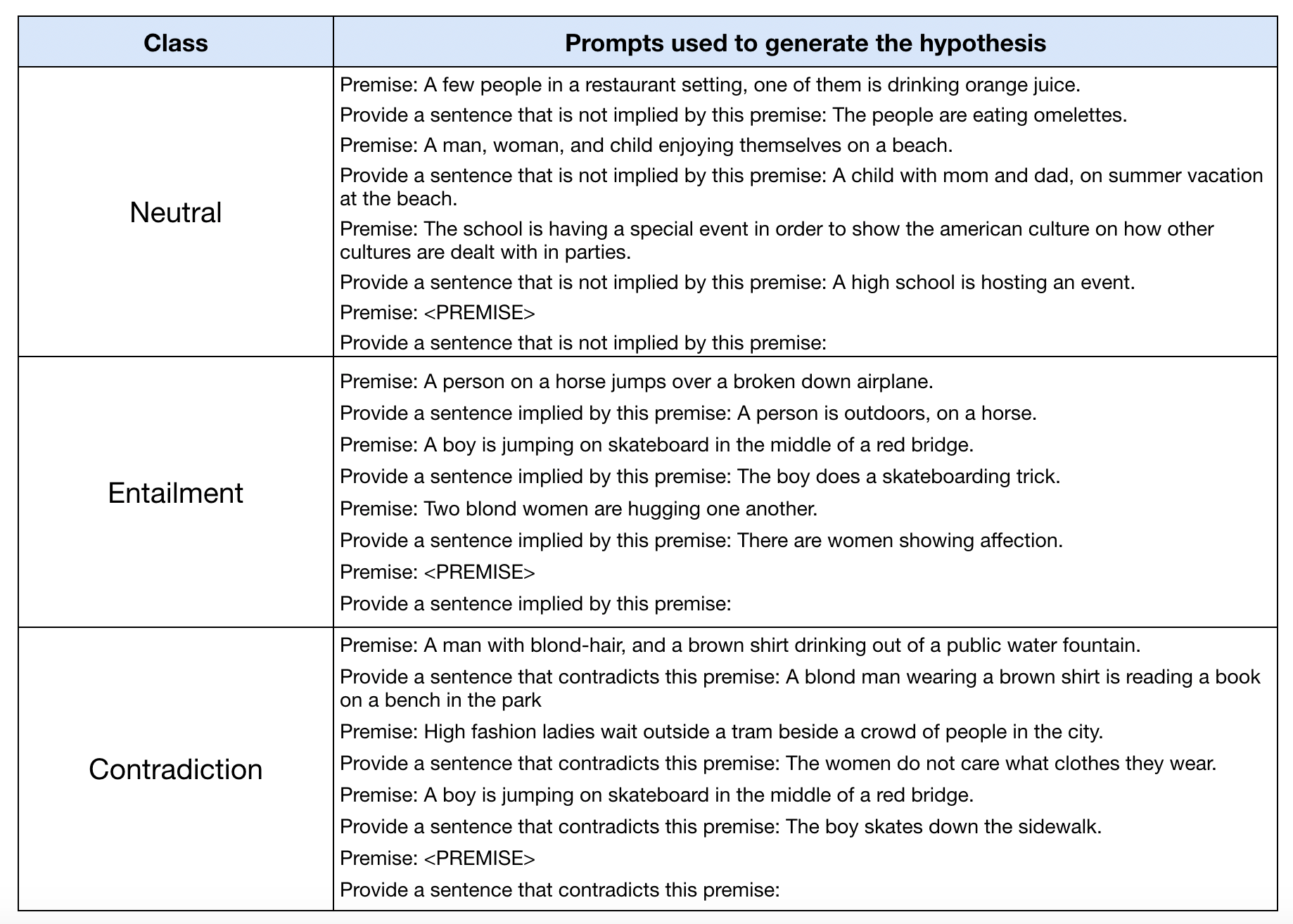} 
    \caption{Prompts used to generate hypotheses for MNLI, where <Premise> contains the premise generated by the generator model.} \label{mnli_prompts}
\end{figure*}

As described in Figure \ref{method_example}, first a prompt is provided to our generator model that asks the generator to create an example extract from a specified domain. For the popular magazine article domain, this prompt asks for an `Example extract from a popular magazine article:', while for the travel guide the prompt is `Example extract from a travel guide:', and for the fiction genre the prompt asks for `Example extract from a fiction book:'. The fourth domain is extracts from government websites, where there are several different subcategories provided in MNLI, either using press releases, letters, speeches or reports. For this fourth domain, we therefore use the following prompts:  `Example extract from a press release on a public domain government website:', `Example extract from a letter on a public domain government website:', `Example extract from a speech on a public domain government website' and `Example extract from a report on a public domain government website:'. The premise generation is zero-shot, with no examples provided to the generator. For the hypothesis generation, three examples from the in-distribution training data are provided. The three examples provided are different depending on the class (see Figure \ref{mnli_prompts}).

\begin{figure*}[ht]
    \includegraphics[width=450pt]{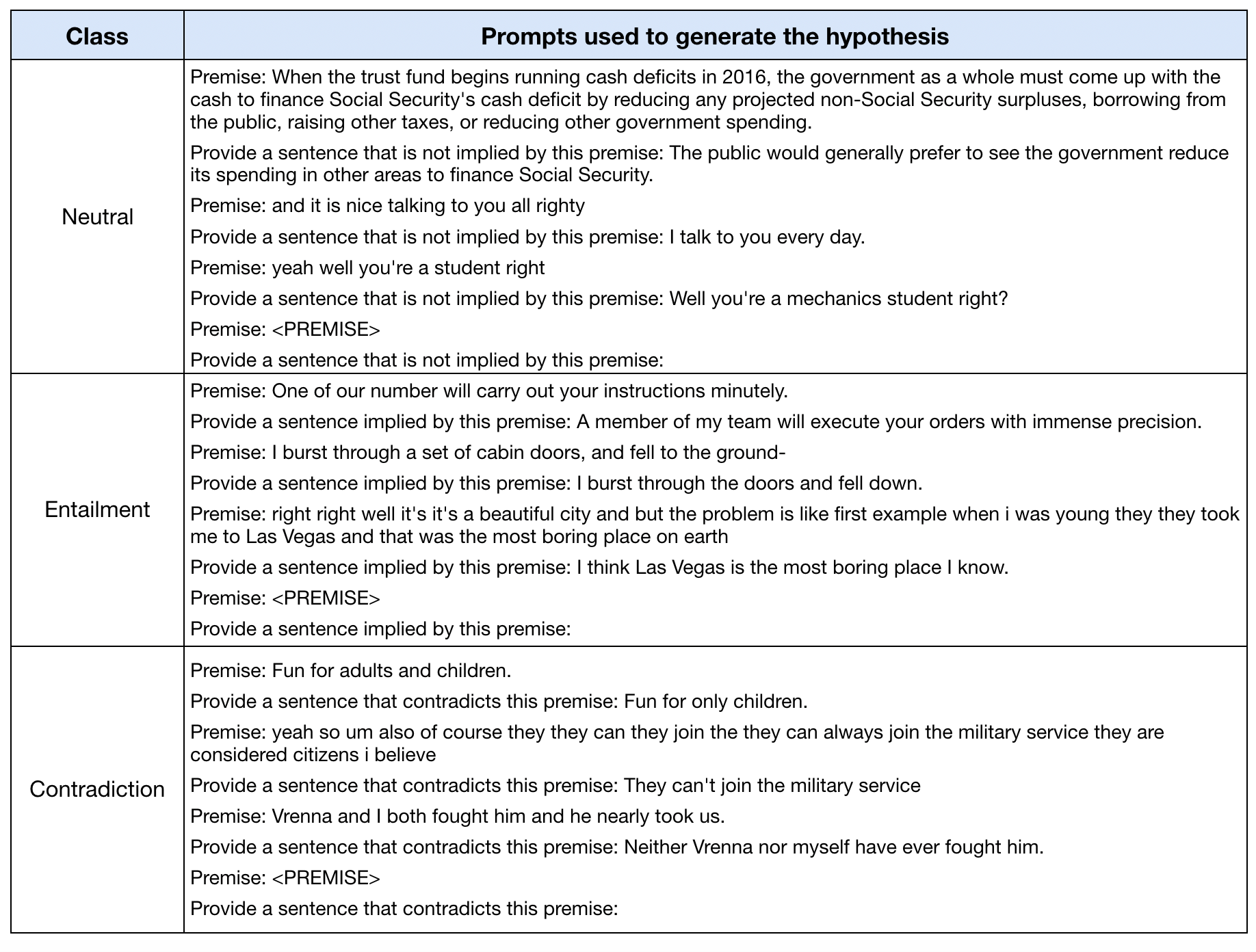} 
    \caption{Prompts used to generate data hypotheses for SNLI, where <Premise> contains the premise generated by the generator model.} \label{snli_prompts}
\end{figure*}

When generating data for SNLI, we generate the the premises using the prompt: `Example flickr image caption:', with the hypotheses generated in the same method described above. As we are using MNLI training data for this setting, the examples in the prompts are provided from the MNLI training data (see Figure \ref{snli_prompts}).

When generating premises for either MNLI or SNLI, only sentences that were at least 8 characters long were included. Premises that finished with a question mark were also not included in the augmented data. If more than one sentence was provided by the generator, and the first sentence did not meet this criteria, then we considered the second sentence as a possible premise.

\begin{figure}[ht]
    \includegraphics[width=\columnwidth]{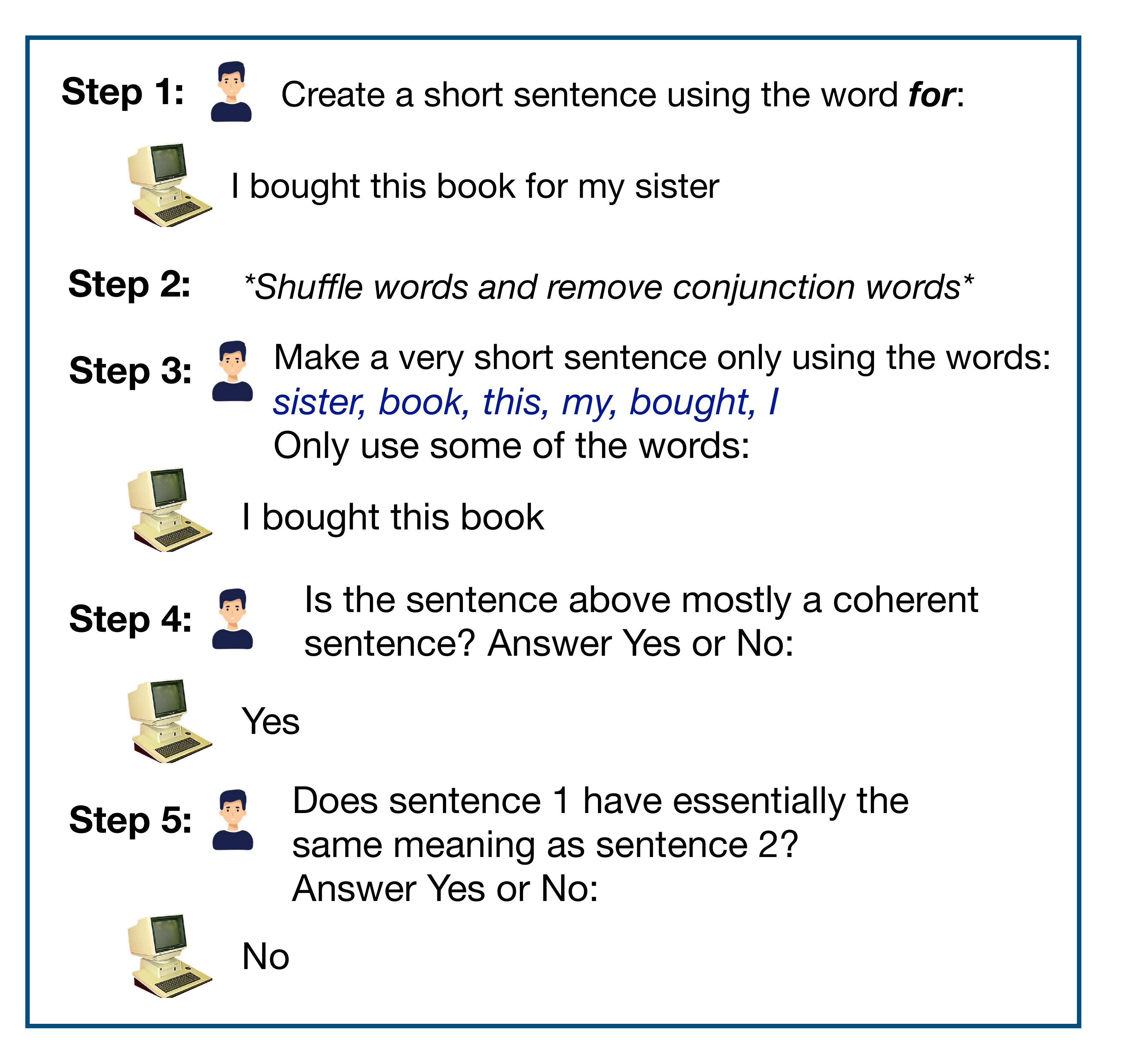} 
    \caption{The process for generating augmented data for our word-overlap augmentation (WOA). In step 4, the model is asked if both the premise and the hypothesis are mostly coherent sentences. In this step, the premise-hypothesis pair is only added to our augmented dataset if the model answers 'yes' for both the premise and the hypothesis. Finally, the sentence pair is only included if the model answers `no' to the final question in step 5.} \label{hans_figure}
\end{figure}

Finally, the prompts used for HANS are provided in Figure \ref{hans_figure}.
\section{Supporting P-values}

In \cref{compression_results_p_vals} we provide the full p-values supporting the statistical testing reported in \cref{compression_results}, \cref{self_distil}, \cref{DeBERTaToBERT},   \cref{compression_results_deberta} and \cref{self_distil_all}. 

\begin{table*}[!t]
\begin{center}
\begin{tabular}{rcccccc}
\toprule
\multirow{2}{*}{\bf Model} & \multicolumn{2}{c}{\bf In-Distribution} & \multicolumn{3}{c}{\bf Out-of-Distribution} \\
\cmidrule(lr){2-3} \cmidrule(lr){4-6}
  &  SNLI-dev & SNLI-test & SNLI-hard & MNLI-mm & MNLI-m \\
\midrule
& \multicolumn{5}{c}{\textit{BERT -> TinyBERT:}} \\
DMU & 0.0796 & 0.0034$\downarrow$ & $<$0.0001$\uparrow$ & $<$0.0001$\downarrow$ & $<$0.0001$\downarrow$ \\ 
DTA & 0.2076 & $<$0.0001$\uparrow$ & 0.0068$\uparrow$ & $<$0.0001$\uparrow$ & $<$0.0001$\uparrow$ \\ 
DTA with DMU & 0.9564 & 0.0042$\uparrow$ & $<$0.0001$\uparrow$ & $<$0.0001 $\uparrow$ & $<$0.0001$\uparrow$ \\ 
\midrule
& \multicolumn{5}{c}{\textit{BERT -> BERT:}} \\ 
DTA & 0.4092 & 0.1580 & 0.0682 & 0.0004$\uparrow$ & 0.0026$\uparrow$ \\ 
DTA (ens) & 0.0846 & 0.2238 & 0.0916 & $<$0.0001$\uparrow$  & $<$0.0001$\uparrow$  \\ 
\midrule
& \multicolumn{5}{c}{\textit{DeBERTa -> DeBERTa:}} \\
DTA & 0.5366 & 0.7934 & 0.6014 & <0.0001$\uparrow$  & $<$0.0001$\uparrow$  \\ 
DTA (ens) & 0.6592 & 0.0570 & 0.2154 & $<$0.0001$\uparrow$ & $<$0.0001$\uparrow$ \\ 
\midrule
& \multicolumn{5}{c}{\textit{DeBERTa -> BERT:}} \\ 
DMU & 0.0102$\uparrow$ & 0.0296$\uparrow$ & 0.0134$\uparrow$ & 0.0896 & 0.5460 \\ 
DTA & 0.0148$\uparrow$ & 0.0372$\uparrow$ & 0.1446 & <0.0001$\uparrow$ &  $<$0.0001$\uparrow$ \\ 
DTA with DMU & 0.0026$\uparrow$ & 0.0148$\uparrow$ & 0.0008$\uparrow$ & $<$0.0001$\uparrow$ & $<$0.0001$\uparrow$ \\ 
\midrule
& \multicolumn{5}{c}{\textit{DeBERTa -> TinyBERT:}} \\ 
DMU & $<$0.0001$\downarrow$ & $<$0.0001$\downarrow$ & $<$0.0001$\uparrow$ & $<$0.0001$\downarrow$ & $<$0.0001$\downarrow$ \\ 
DTA & 0.7730 & <0.0001$\uparrow$ & 0.0238$\uparrow$ & $<$0.0001$\uparrow$ & <0.0001$\uparrow$ \\ 
DTA with DMU & $<$0.0001$\downarrow$ & $<$0.0001$\downarrow$ & $<$0.0001$\uparrow$ & $<$0.0001$\uparrow$ & $<$0.0001$\uparrow$ \\ 
\bottomrule
\end{tabular}

\end{center}
\caption{P-values for our main results tables, comparing our methods to standard knowledge distillation. We use two-tailed bootstrapping hypothesis testing \citep{efron1993introduction} to test statistical significance. $\uparrow$ represents results where there is a significant improvement compared to the baseline, whereas $\downarrow$ represents results that are significantly worse than the baseline. For the BERT -> BERT and DeBERTa -> DeBERTa settings, we compare our DTA method to standard knowledge distillation, while our DTA (ens) method is compared to standard knowledge distillation using an ensemble of teacher models. }
\label{compression_results_p_vals}
\end{table*}

\end{document}